# Improving Global Weather and Ocean Wave Forecast with Large Artificial Intelligence Models

**Authors:** Fenghua Ling[1,2†], Lin Ouyang[1†], Boufeniza Redouane Larbi[1†], Jing-Jia Luo[1*], Tao Han[2,3], Xiaohui Zhong[4], Lei Bai[2*]

**Affiliations:**
[1]Institute for Climate and Application Research (ICAR)/CIC-FEMD/KLME/ILCEC, Nanjing University of Information Science and Technology, Nanjing, China
[2]Shanghai AI Laboratory, Shanghai, China
[3]The Hongkong University of Science and Technology, Hongkong, China
[4]Fudan University, Shanghai, China

†Equal Contributions,
*Corresponding to jjluo@nuist.edu.cn; baisanshi@gmail.com

**ABSTRACT**

The rapid advancement of artificial intelligence technologies, particularly in recent years, has led to the emergence of several large parameter artificial intelligence weather forecast models. These models represent a significant breakthrough, overcoming the limitations of traditional numerical weather prediction models and indicating the emergence of profound potential tools for atmosphere-ocean forecasts. This study explores the evolution of these advanced artificial intelligence forecast models, and based on the identified commonalities, proposes the "Three Large Rules" to measure their development. We discuss the potential of artificial intelligence in revolutionizing numerical weather prediction, and briefly outlining the underlying reasons for its great potential. While acknowledging the high accuracy, computational efficiency, and ease of deployment of large artificial intelligence forecast models, we also emphasize the irreplaceable values of traditional numerical forecasts and explore the challenges in the future development of large-scale artificial intelligence atmosphere-ocean forecast models. We believe that the optimal future of atmosphere-ocean weather forecast lies in achieving a seamless integration of artificial intelligence and traditional numerical models. Such a synthesis is anticipated to offer a more advanced and reliable approach for improved atmosphere-ocean forecasts. Additionally, we illustrate how forecasters can adapt and leverage the advanced artificial intelligence model through an example by building a large artificial intelligence model for global ocean wave forecast.

**1. INTRODUCTION**

In the development of meteorology, weather forecasting stands out as one of the most crucial and extensively scrutinized branches, influencing various social and economic decisions (Alley et al. 2019). Particularly under global warming, the increased frequency of extreme events (Rahmstorf and Coumou 2011) amplifies the significance of accurate weather forecasts. The accuracy of the forecasts is critical for effective emergency management, mitigation of economic losses, and facilitation of revenue generation across pivotal sectors such as energy, agriculture, and transportation (Bauer et al. 2015; Lazo et al. 2009).

In the early 20$^{th}$ century, Abbe and Bjerknes (Abbe 1901; Bjerknes 1904) laid the foundation for modern numerical forecasting by proposing physical laws to predict weather, framing atmospheric state prediction as an initial value problem in mathematical physics. The advent of supercomputers and advanced observational instruments, facilitated the solution of complex partial differential equations, marking a critical shift in weather forecast. Consequently, weather forecasting transitioned from reliance on empirical reasoning to the era of numerical weather prediction (NWP, Charney et al. 1950; Lynch 2008). Over the past fifty years, NWP has undergone significant transformation through the integration of data assimilation (Charney et al. 1950b; Courtier et al. 1994) and physical parameterization (Stensrud 2009; Williams 2005). This integration has quietly revolutionized NWP, substantially enhancing prediction capabilities. Nevertheless, global medium-range weather forecast still faces significant challenges due to uncertainties in initial and boundary conditions, the complexity of nonlinear physical processes, and the substantial computational demands (Benjamin et al. 2018). As a result, the skills of medium-range weather forecast remain constrained, with reliable prediction being limited to approximately nine days (Alley et al. 2019; Bauer et al. 2015; Benjamin et al. 2018).

The field of weather forecast is currently experiencing a profound transformation, driven by advances in computing science and the application of artificial intelligence (AI) methods in weather forecasting (Dueben and Bauer 2018; Rasp et al. 2020; Scher 2018; Schultz et al. 2021; Weyn et al. 2019). Recently, the emergence of Large AI Weather forecast Models (LWM), such as FourCastNet (Pathak et al. 2022), Pangu-Weather (Bi et al. 2023), GraphCast (Lam et al. 2023), FengWu (Chen Kang et al. 2023), and FuXi (Chen L. et al. 2023a), showcases the potential of AI-driven methods to surpass traditional dynamical models and extending the current weather forecast limit (Chen Kang et al. 2023; Chen L. et al. 2023a; Rasp et al. 2023). Significantly, the European Center for Medium-Range Weather Forecasts (ECMWF), one of the world's top meteorological agencies, has also begun adopting AI technology for weather forecasts (AIFS, https://www.ecmwf.int/en/about/media-centre/aifs-blog/2023/ECMWF-unveils-alpha-version-of-new-ML-model). These AI models not only offer more accurate predictions but also largely reduces computational costs (Bi et al. 2023; Chen Kang et al. 2023;

Chen L. et al. 2023a; Lam et al. 2023; Pathak et al. 2022; Rasp et al. 2023). The impact of AI methods on meteorological service is comparable to the revolutionary changes brought about by AlphaFold in medicine (Jumper et al. 2021). This leads to a critical question: are we witnessing a paradigm shift in weather forecasting?

This study aims to provide a relatively comprehensive overview of the current state-of-the-art LWMs (see Table 1), analyzing and summarizing their capabilities and limitations. Furthermore, we investigate the potential challenges that LWMs may face in weather forecasting. One goal is to enhance collaborations between "data scientists" and "weather forecasters", thereby contributing to the development of more effective hybrid approaches that integrate data-driven and physics-based methods for better meeting the requirements in atmosphere and ocean sciences.

Table 1. The representative LWMs developed since 2020

| Model | Institution | Spatial-Resolution | Methods | First publication time |
|---|---|---|---|---|
| CNN (WeatherBench) | Technical University of Munich | 5.625° | CNN-based | 2020.08 |
| Resnet (WeatherBench) | Technical University of Munich | 5.625° | CNN-based | 2021.08 |
| FourCastNet | NVIDIA | 0.25° | Transformer-based | 2022.02 |
| SwinVRNN | Alibaba Damo Academy | 5.625° | Transformer-based | 2022.05 |
| Pangu-Weather | Huawei Cloud | 0.25° | Transformer-based | 2022.11 |
| GraphCast | Google DeepMind | 0.25° | GNN-based | 2022.12 |
| ClimaX | Microsoft | 1.4° | Transformer-based | 2023.01 |
| FengWu | Shanghai AI Lab | 0.25° | Transformer-based | 2023.04 |
| FuXi | Fudan University | 0.25° | Transformer-based | 2023.06 |
| ACE | Allen Institute for AI | 1° | Transformer-based | 2023.09 |
| AIFS | ECMWF | 1° | GNN-based | 2023.10 |
| Neural GCM | Google | 0.7° | Hybrid model | 2023.11 |
| GenCast | Google | 1° | Diffusion model | 2023.12 |
| FengWu-GHR | Shanghai AI Lab | 0.09° | Transformer-based | 2024.01 |

## 2. Several advanced LWMs

In the past decades, simple AI methods, such as support vector machine, random forest, neural network, were explored for solving various tasks including curve fitting, linear regression, or data assimilation. However, performance of these methods is often limited (Azimi-Sadjadi and Zekavat 2000; Hsieh et al. 1998). With the advancement of deep learning algorithms, an increasing number of interdisciplinary studies have adopted this technology into weather forecasting. In this section, we highlight representative endeavors in the application of advancec deep learning methods to global medium-range weather forecasting since 2020, focusing on notable models such as FourCastNet, Pangu-Weather, GraphCast, FengWu, and FuXi, which have caused significant attentions recently. We will delve into the definition and significance of large parameter AI models in the field of meteorology.

The pioneering work in medium-range weather prediction utilizing artificial intelligence was conducted by Dueben and Bauer (Dueben and Bauer, 2018). They introduced a toy model based on neural networks and demonstrated that AI models are competitive with forecasts produced by a complex atmosphere general circulation model with a coarse resolution (TL21). Subsequently, Convolution Neural Network (CNN, Rasp et al., 2020) and Residual neural Network (ResNet, Rasp et al., 2021) were employed to generate forecasts for global 500 hPa geopotential height and 850 hPa air temperature at a spatial resolution of 5.625°. Although these AI methods did not outperform the ECMWF Integrated Forecast System (IFS) model's High-Resolution deterministic forecast (HRES), they established a comprehensive dataset and framework of training and evaluation for AI weather forecasts, aptly named WeatherBench (Rasp et al. 2020). This pioneering work serves as a reliable and rational benchmark for subsequent studies in this field.

Despite the increasing efforts following the establishment of WeatherBench, most of early studies initially concluded that while the AI application is valuable, practical implementation seemed to be improbable (Dueben and Bauer 2018; Scher 2018; Weyn et al. 2019). However, this challenge was tackled by Pathak et al. (2022); they applied an improved vision transformer called FourCastNet, in which AFNO (Adaptive Fourier Neural Operator, Guibas et al. 2021) provides the underlying architecture for global weather forecasting. AFNO divides the input frame into blocks or markers and uses fast fourier transform spatial blending operations, which helps capture fine-scale phenomena and reduce errors in coarse-grid models, resulting in 0.25-degree resolution and 6-hourly outputs, using a rolling forecast method to predict weather during next 7 days with extremely high efficiency. It was found that skill of the AI method is close to the current best dynamical model forecasts of the first 3-day weather. More importantly, forecasts using the AI model can be accomplished by only one GPU within one minute, which consumes much less energy than that required by IFS.

More and more large technology companies and research institutions have discovered that AI has great potential for providing better NWP service (Hu et al. 2023; Nguyen et al. 2023). In this technology competition, Huawei Cloud's Pangu-Weather took the lead. They drew on the architecture of Swin Transformer (Liu et al. 2021), considered more vertical layers (13 levels) than FourCastNet, and regarded height information as a new dimension. A 3D Earth Specific Transformer (3DEST) was designed by encoding geographical features into the network, making it easier for the model to capture the relationship among the atmospheric states at different pressure levels (Bi et al. 2023). The hindcast experimental results show that Pangu-Weather trained with ERA5 outperforms IFS-HRES for the first time at a high spatiotemporal resolution. And the forecast of extreme events such as typhoon tracks is also better than that of HRES.

While Transformer-based models have exhibited commendable capabilities, the Google team emphasizes the importance of aligning modeling approaches with the Earth's spherical nature and incorporates additional physical considerations. As a result, they embarked on exploring a graph neural network-based model architecture (Keisler 2022; Pfaff et al. 2020), leading to the inception of GraphCast (Lam et al. 2023). This innovative model significantly outperformed IFS-HRES on over 90% of 1380 predictands with the effective forecast time (with ACC of Z500 prediction being greater than 0.6) is 9.25 days (Alley et al. 2019; Bauer et al. 2015). A noteworthy aspect of GraphCast lies in its introduction of "multi-mesh" representation modeling. From a data science perspective, the overlay of conventional hexagonal grid levels enhances the model's coding process, allowing for the inclusion of both short-range and long-range connections, thereby facilitating more efficient information dissemination. The incorporation of "multi-mesh" representation in GraphCast not only enhances its performance but also serves as an innovative approach that bridges the realms of data science and meteorology.

With these advances, it is curious for the community to explore whether AI methods can push the upper bound of NWP and be effective in operational forecast setting. In a groundbreaking development, FengWu (Chen Kang et al. 2023), proposed in April 2023 by Shanghai Artificial Intelligence Lab, has extended the skillful weather forecasts up to 10.75 days lead for the first time. Notably, FengWu explores the implementation and deployment using the same operational analysis as IFS-HRES. It has been verified that FengWu can operate as a pseudo-real-time prediction system, maintains its leading position. The model's robust performance showcases its potential not only in extending the lead times of skillful weather predictions but also in operational real-time forecast applications.

The success of FengWu can be attributed to its better mining of atmospheric data and the application of probabilistic method, which defines weather forecasting as a probabilistic task. A crucial aspect of FengWu is its treatment of each variable as an independent modality and the utilization of a multi-modal transformer architecture. This approach enables the effective modelling of complex intra- and inter-correlations among the high-dimensional atmospheric data. In contrast, the Pangu model treats all variables as a single modality. Recently, Shanghai Artificial Intelligence Lab proposed FengWu-GHR to further improve the accuracy of weather forecasts by employing transfer learning strategies and model structure design to increase the spatial resolution of medium-range weather forecasts to 9km, extending the skillful weather forecasts up to 11 days (Han et al. 2024).

Similarly, FuXi AI model, proposed by Fudan University, distinguished itself by using an innovative cascade approach that integrates three pre-trained U-Transformer models to reduce the cumulative error of AI-based weather forecast (Chen L. et al. 2023a). The three models are meticulously fine-tuned for optimized forecast performance at lead times of 0-5 days, 5-10 days, and 10-15 days, respectively. This fine-tuning process ensures that each model excels within its specified time window. A key innovation of FuXi lies in the cascade of these models to generate a comprehensive 15-day forecast. This methodology significantly enhances the accuracy and reliability of medium-range weather forecasts. In addition, FuXi incorporates forecasts for diagnostic quantities such as total precipitation, a notable addition compared to other models. Leveraging on this framework, the Fudan University team has conducted extensive studies across various forecast timescales and developed a suite of models, including FuXi-extreme and FuXi-S2S (Zhong et al. 2023; Chen L. et al. 2023b).

While the aforementioned models have been highlighted for their excellence, several others also deserved attention, including those that combine external forcing to predict weather (ACE, Watt-Meyer et al. 2023), physical-informed hybrid models (Neural GCM, Kochkov et al. 2023), probabilistic forecasting model (GenCast, Price et al. 2023) and AIFS with 1° spatial resolution proposed by ECWMF itself, among others. While these noteworthy models are not individually discussed in this article, their basic information is listed in Table 1 for reference.

With the increasing number of LWM, there is an urgent need to propose a standardized definition of LWM. We conducted a comparison of four models (Table 2): Pangu, GraphCast, FengWu, and FuXi. Basedon the common characteristics shown in Table 2, we propose the following three criteria, termed the "Three Large Rules":

**Large Parameter Count:** These models are characterized by their large number of parameters or significant computational complexity. Typically, these models encompass tens of millions to even billions of parameters. Such a large parameter count enhances the model's proficiency in capturing intricate features and patterns, thereby improving its adaptability to a wide range of input data. While an excessively high parameter number may pose training challenges and potential overfitting, a model with a small number of parameters (like Resnet or other toy models, Dueben and Bauer 2018; Scher 2018; Weyn et al. 2019) may not be categorized as an LWM.

**Large Number of Predictands:** These models excel in generating forecasts for a wide variety of meteorological variables with a high spatial resolution and at different pressure levels, offering detailed information on the atmospheric vertical structure and surface conditions.

**Large Scalability and Downstream Applicability:** LWMs exhibit remarkable scalability, enabling them to extend beyond basic weather forecast tasks. It enables the development of specialized models for a wide range of downstream tasks. For example, the LWM's model structure can be utilized to explore AI methods for data assimilation, allowing for the integration of observational data (Xiao et al. 2023; Chen Kun et al. 2023). The extracted atmospheric data features from the LWM can be applied to downstream tasks such as downscaling and precipitation forecasting (Nguyen et al. 2023; Pathak et al. 2022). Furthermore, the predictions of the LWM can be utilized in the construction of decision models like energy scheduling. This scalability defines their significance in effectively addressing a broad range of meteorological and oceanic services. This scalability defines their significance in effectively addressing a broad range of meteorological services. Similarly, Global Climate Models (GCMs) seem to satisfy these three large rules. Due to the chaotic nature of the atmosphere, GCMs require numerous parameterization processes to accurately simulate the overall atmospheric variations. Additionally, GCMs can handle large amounts of input data, generate diverse prediction and also exhibit a high scalability. The key difference between GCM and LWMs lies in the fact that GCMs are built based on dynamical and thermodynamical equations, while LWMs rely on data to construct statistical mapping relationships, gradually approaching the functional capabilities of GCMs. Therefore, we believe that AI models meeting the proposed "Three Large Rules" or approaching GCMs' capabilities can be referred to as LWMs.

Table 2. The comparison of the functionalities of Pangu, GraphCast, FengWu, and FuXi weather forecast models. *Indicates the results reproduced in this study, which may be somewhat different from their original reports.

| Model | Number of predictands | Resolution (all have 0.25° spatial resolution) | Parameters | Computational complexity* |
|---|---|---|---|---|
| Pangu-Weather | 69 | Vertical: 13 pressure level and surface Frequency: 1h, 3h, 6h, 24h | 740 million | 4950 Gflops |
| GraphCast | 227 | Vertical: 37 pressure level and surface Frequency: 6h | 55 million | 13564 Gflops |
| FengWu | 189 | Vertical: 37 pressure level and surface Frequency: 6h | 751 million | 8000 Gflops |
| FuXi | 70 | Vertical: 13 pressure level and surface Frequency: 6h | 1556 million | 19892 Gflops |

## 3. Why did LWM change NWP?

As discussed above, the LWMs have been increasingly integrated into mainstrem medium-range weather forecasts in recent years. This section analyzes the pivotal factors contributing to the revolutionary impact of AI on weather forecast.

**Accuracy**: The AI-based models demonstrate a remarkable improvement in forecast accuracy compared to traditional physical models. This is evident in the comparison illustrated in Figure 1, which presents the skills in forecasting weather in 2018 based on Pangu-Weather, GraphCast, FengWu and FuXi models and ECMWF dynamical models. Skill assessments for Z500 and T850 prediction were conducted following the methodology outlined in WeatherBench2 (Rasp et al. 2023). The results are conclusive: AI models significantly outperform the IFS, particularly in predicting extreme weather events, like typhoon tracks, atmospheric rivers, extreme high temperatures, etc. Moreover, the rapid development of AI techniques suggest that, with improved computing resources and innovative methlogies, AI models have great potential to further surpass dynamical models for both deterministic and ensemble forecasts.

The superior performance of AI-based models over traditional models can be attributed to several key factors. Firstly, AI-based models excel in using historical data for iterative refinement. It is well known that dynamical models produce long-lead predictions by iteratively integrating partial differential equations, leading to substantial error accumulation, particularly in a strong nonlinear system. In contrast, AI-based models employ various strategies like cascade (Fuxi) and replay buffer mechanisms (FengWu), which are specifically fine-tuned to help effectively mitigate error accumulation, a critical advantage for long-lead predictions. Secondly, dynamical models often rely on a large number of oversimplified formulae, physical parameterizations, and empirical equations, resulting in significant deviations from the truth during forecasts. In contrast, contemporary large AI models exhibit robust fitting and extrapolation capabilities. For instance, AI video prediction techniques are able to derive non-linear mappings from historical and future data, tackling the interactions across different timescales and spaces. This approach significantly enhances weather forecast performance, overcoming the limitations of empirical formulae.

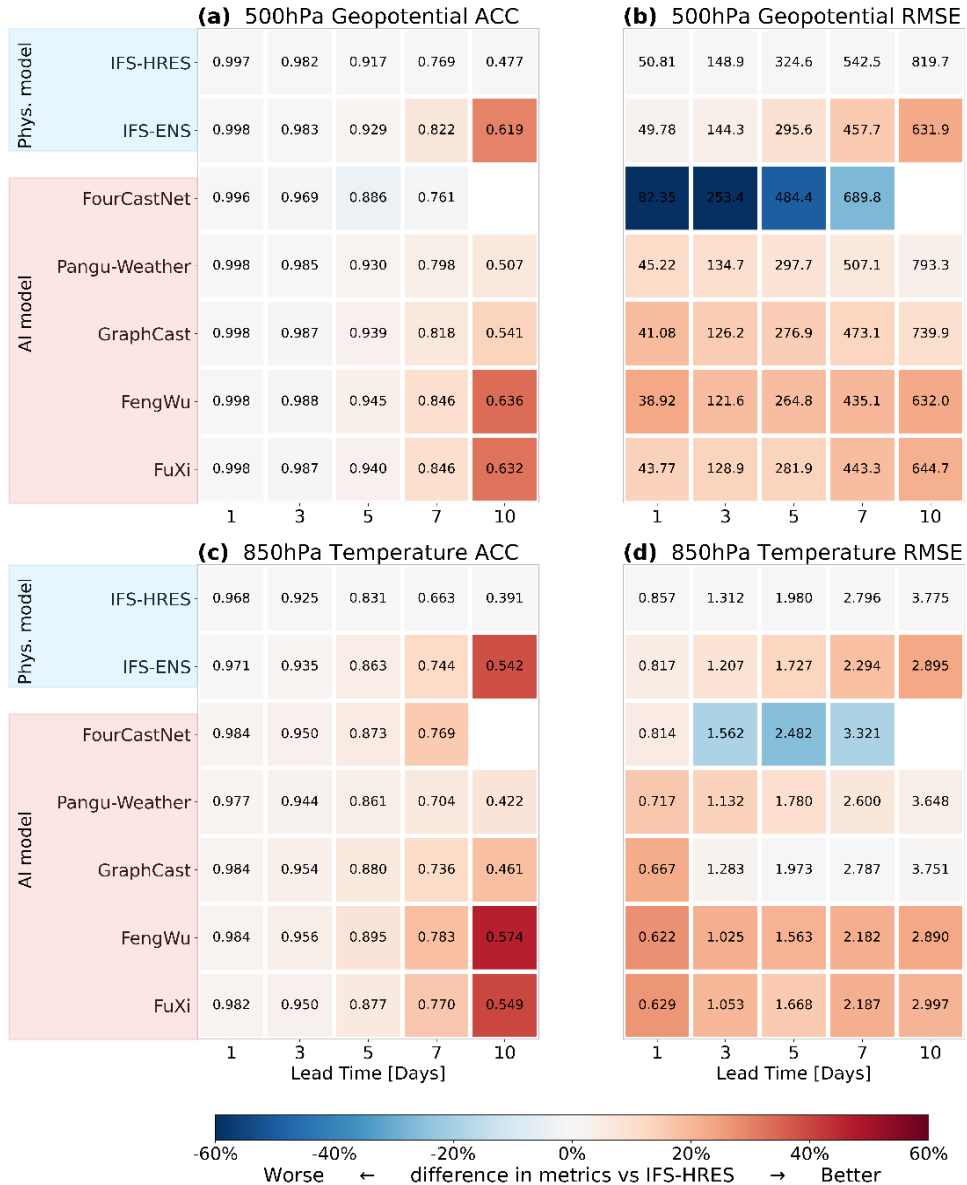

**Figure 1.** Deterministic headline scorecards for geopotential height at 500 hPa (a and c) and temperature at 850 hPa (b and d). Values show root mean square error (RMSE) and anomaly correlation coefficient (ACC). Colors denote the difference (%) to the IFS HRES baseline (Adapted from Rasp et al. 2023).

**Efficiency**: Traditional NWP, such as the IFS forecast, requires substantial computation resources. For instance, a 15-day ensemble forecast with 51 members at a spatial resolution of 18km, running on 1530 Cray XC40 nodes, takes a considerable 82 minutes (Bauer et al., 2020). This illustrates the resource-intensive nature of conventional NWP approaches. In contrast, AI-based large models demonstrate remarkable efficiency, particularly when utilizing GPUs or TPUs. Even the slowest and most computationally demanding model, GraphCast, can produce a 10-day forecast within 60 seconds using only one GPU.

The extremely low computational cost of AI-based weather forecast models opens up a new avenue for constructing high-resolution ensembles with potentially hundreds of members, a significant leap from the traditional 50-member ensembles. Furthermore, the user-friendly deployment and open-source nature of these AI models enable individuals to perform customized weather forecasts on personal computers. This facilitates the dissemination of information beyond large opertional centers. Users can now easily generate 7-15 days weather forecasts by accessing relevant initial field data. This paradigm-shift highlights the revolutionary impact of AI on weather forecasts, making it more accessible and efficient.

**4. AI-based LWM still poses significant challenges**

While LWMs hold great promise, the journey to future development is riddled with challenges. In this section, we will delve into a discussion of the obstacles that AI models must surmount throughout the entire NWP process to achieve an end-to-end forecasting model.

**Data quality control and assimilation**: Undoubtedly, weather forecast performance depends on the accuracy of initial values. Currently, many AI-based models rely on the initial fields provided by operational systems like IFS. It is crucial for these models to reduce their dependence on dynamical model analysis data and prioritize real-time data input in practical scenarios. Quality control becomes highly important for real-time data input, given the existence of various errors or deficiencies in satellite, buoy, and station/gauge observations. AI methods, with their advantages in quality control and data reconstruction, offer a promising avenue for addressing these issues (Neukom et al. 2019; Tsagkatakis et al. 2019; Wang et al. 2022).

AI methods must explore more sophisticated strategies for effectively leveraging real-time data input. Currently, there exist two predominant paradigms in this realm. The first involves the direct utilization of sparse data or station data as predictors, treating it as a distinct modality during AI model design. Examples of this approach include Corrformer (Wu et al. 2023), Met-Net3 (Andrychowicz et al. 2023) and DeepPhysiNet (Li et al. 2024), which were primarily applied to regional weather forecasts. However, endeavors to implement this paradigm on a global scale require further in-depth exploration. The second paradigm seeks to emulate the conventional dynamical model approach by introducing an assimilation module in AI-based models. For example, Melinc and Zaplotnik (2023) used traditional 3-Dimensional variational with AI methods to perform data assimilation in the latent variable space, saving much of the computing resources. Chen Kun et al. (2023) conducted an assimilation experiment based on FengWu by building an AI model for data assimilation (FengWu-Adas), demonstrating that the LWM does have end-to-end potential. Xiao et al. (2023) proposed Fengwu-4Dvar, which utilizes automatic derivation in the AI framework to emulate the adjoint model of 4-dimensional variational assimilation, enabling the LWM to generate good analysis data for accurate and efficient iterative predictions.

**Ensemble prediction**: Even with the most advanced data assimilation systems currently available, it is still impossible to create an accurate initial field. Therefore, it is crucial for a forecast system to account for the uncertainties in initial conditions. Generating a multitude of possible initial fields and producing rational probabilistic forecasts become necessary to provide additional guidance for decision-making. The belief in the potential of AI models to enhance the accuracy of weather forecasts with large ensemble members is widespread. However, a critical challenge lies in designing perturbation schemes of initial state aligned with the characteristics of AI models and the physical laws. This intricate task requires strategic solutions to ensure the reliability and effectiveness of ensemble forecast approaches.

**Physical information-guided neural network design**: Currently, it is still controversial whether the LWM can learn physics and have physical expression. Seltz and Craig (2023) contended that current AI-based models, exemplified by Pangu Weather, fall short in simulating the butterfly effect, incorrectly implying infinite atmospheric predictability. This limitation suggests that LWM ensembles may struggle to mitigate sampling uncertainty, hindering the provision of more reliable predictions of extreme event occurrence. Ben-Bouallegue et al. (2023) asserted that current LWMs encounter similar prediction challenges as dynamical models like IFS, indicating that AI models are still bound by certain physical constraints. Hakim and Masanam (2023) examined Pangu and found that the AI model possesses specific physical properties, making it suitable for numerical sensitive experiments.

These debates highlight the need for future AI weather forecast models to prioritize not only the prediction accuracy but also the correct expression of physical principles, such as the continuity equation in the AI models (Zhang et al. 2023); this is to explore innovative approaches such as building AI and physics-fused hybrid models. Recently, Google team has made a great progress in this direction. They built Neural GCM with two modules: a differentiable dynamic core for solving discretized Navier-Stokes equations, and a neural network for simulating physical parameterization such as cloud formation, radiative transfer and precipitation. Through the combined action of the two modules, Neural GCM can perform medium-range weather forecasts and multidecadal climate simulations (Kochkov et al. 2023). Hybrid atmosphere forecast models, which amalgamate the strengths of AI and traditional physical principles, are paving a way for the future development of weather forecasts and climate simulations.

**Post-processing of prediction results**: Although current AI methods start to outperform IFS in medium-range forecasts of conventional weather variables, it is necessary to expand the predictions for precipitation, visibility, wind and solar energy, and so on, which are challenging but of significant importance for social sustainability. In addition, the majority of current AI models operate at a spatial resolution of 0.25º with 6-hourly outputs. The integration of super-resolution technology proves to be beneficial for refining regional predictions (Ling et al. 2024; Pan et al. 2023; Liu et al. 2024). In the pursuit of continuous improvement in weather forecasts, a crucial strategy involves implementing bias correction and downscaling of prediction results through the development of secondary models (Hess et al. 2022; Ling et al. 2022). This approach contributes to improve accuracy and reliability of weather forecasts, enhancing the usefulness of LWM forecasts for specific tasks.

**5. Global ocean wave forecasts with an LWM**

Indeed, building LWMs require significant computational and storage resources, making it challenging for individuals to train their own LWMs. However, leveraging the predictions from LWMs allow for more in-depth exploration and specific applications. For example, AI large models can be utilized to improve ensemble forecasts, to design sensitivity experiments for scientific discoveries, and to construct bias correction or downstream task models. This section presents an attempt to integrate global ocean wave forecasts with LWM.

The model's training and evaluation are conducted based on the ERA5 reanalysis dataset with 1 resolution. Ocean waves are characterized by three components: significant wave height (SWH), mean wave period (MWP), and mean wave direction (MWD). The meridional and zonal winds at 10-meter high are adopted as the major external forcing, and the depth data from the National Oceanic and Atmospheric Administration (NOAA)'s Earth TOPOgraphy dataset (ETOPO) of the United States was used to fix the terrain boundary.

The structural design of the LWMs is mostly based on the Transformer structure proposed by Vaswani et al. (2017), its advantage over recurrent neural network (RNN) consists of its ability to fully rely on attention mechanisms, eliminating recursion and convolution, and thus leveraging the advantages of parallel computing. Consequently, this structure is also adopted for building the ocean wave prediction model in this section.

In the encoding stage, the Vision Transformer (ViT) proposed by Dosovitskiy et al. (2020) is used to cut the two-dimensional image into many patches and map them to vectors, thus participating in subsequent Transformer Encoder modules. The structure of the ViT is shown in Figure 2. The input comprises initial wave data, future wind information, and terrain details. Given that the terrain information is treated as a static variable, terrain encoding is employed instead of ViT position encoding. Subsequently, these inputs undergo two sequential attention blocks to execute temporal and spatial attention among different time steps and spatial patches. Residual connections and layer normalization are applied to each attention block. Following multiple temporal and spatial attention modules, it can be asserted that the spatiotemporal characteristics of ocean waves have been extracted through self-attention. As waves are strongly driven by winds, the future wind input serves as the decoder input. Cross-attention is implemented between the wind information and the wave information from the encoder stage to articulate the interaction between winds and waves, specifically, how wind drives the development of waves. After several blocks, several convolutional layers are applied to address the fragmentation phenomenon caused by the model. The final predicted waves of the model are obtained after this process.

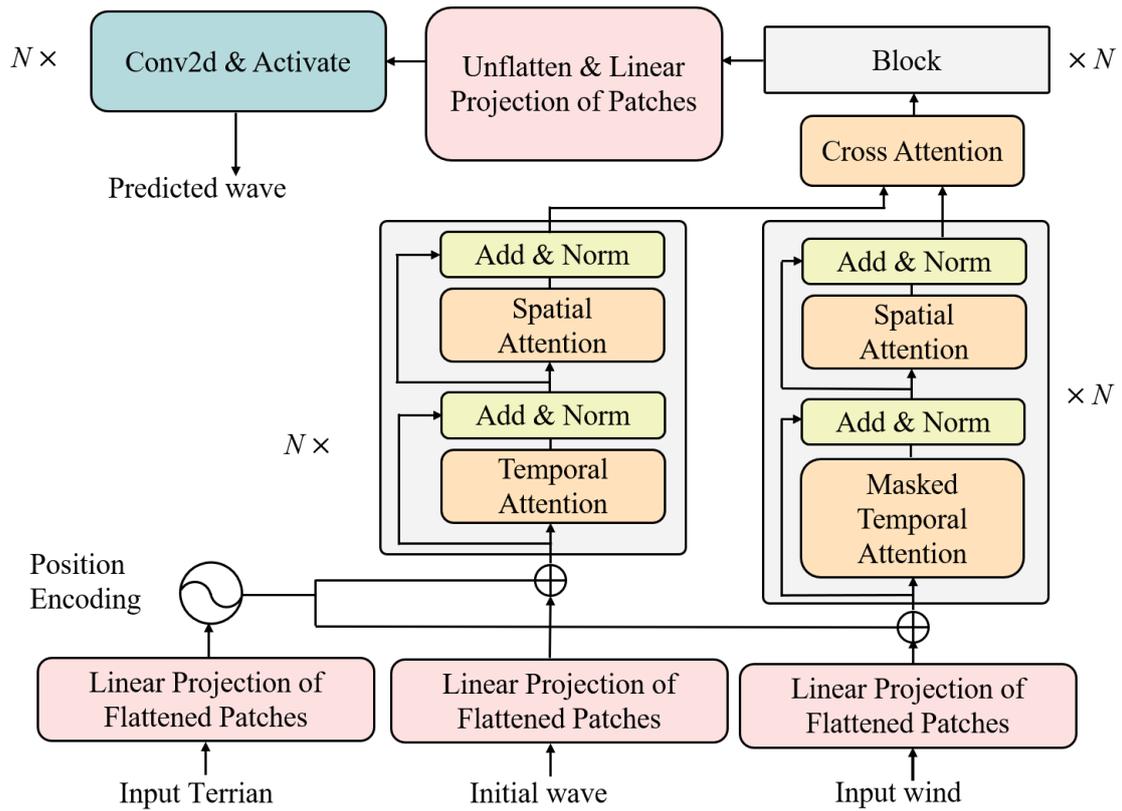

**Figure 2.** The vision transformer model architecture

Figure 3 shows the spatial distribution of RMSE for ocean waves predicted by the ViT model. It shows small errors in tropical regions at lead times of up to 7 days, and large errors appear in mid-high latitudes of the both hemispheres. In the North Pacific and North Atlantic regions, the RMSE of wave height, period and direction forecasts exceeds 1 meter, 2 seconds, and 45°, respectively.

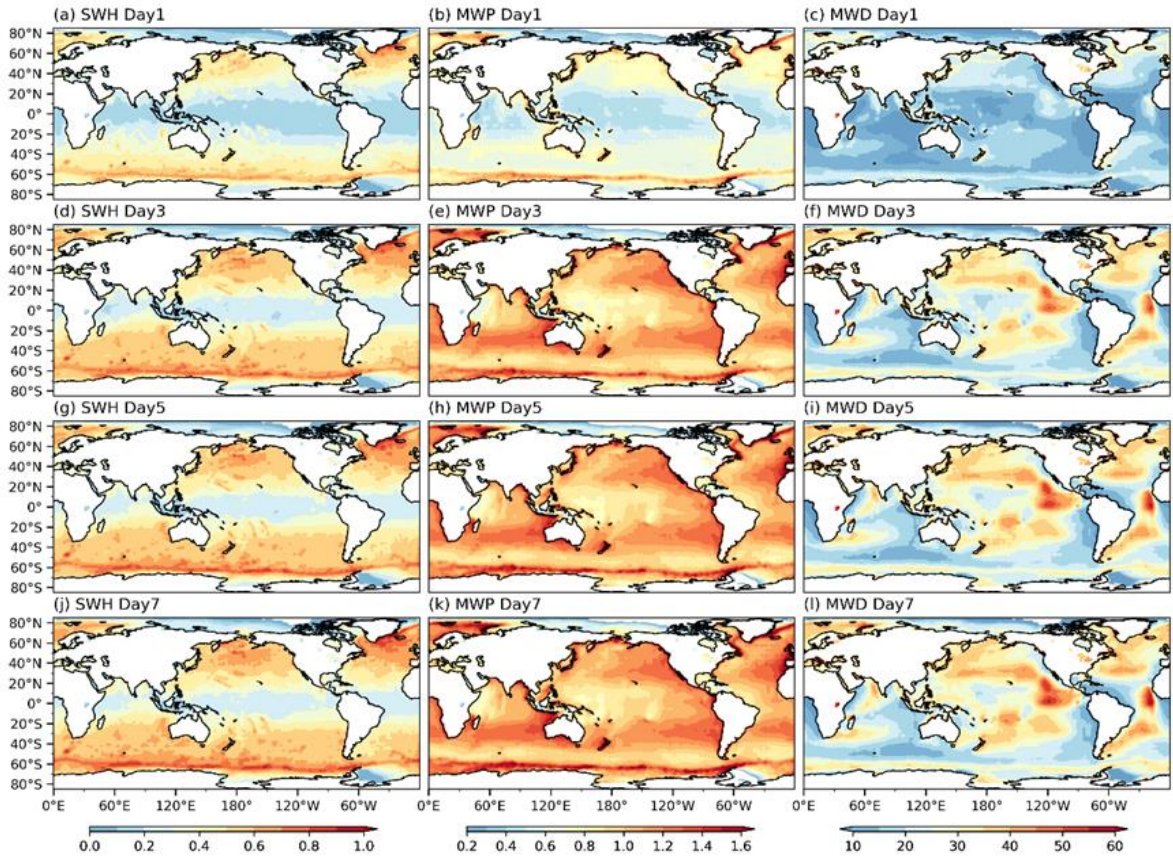

**Figure 3.** The RMSE of (a, d, g, j) SWH, (b, e, h, k) MWP, and (c, f, i, l) MWD predicted by the Vision Transformer model at lead times of 1, 3, 5 and 7 days.

Certain wave heights can be destructive, and the precision in forecasting waves of a specific height is essential. Figure 4 displays the mean relative error (MRE) of wave forecasts with waves of less than 1 meter high being filtered out. Except for polar regions, relative errors of the predicted waves stay at around 5% of their magnitudes at 1-day lead, and the errors increase to about only 10% at lead times up to a week, which still remains within an acceptable range.

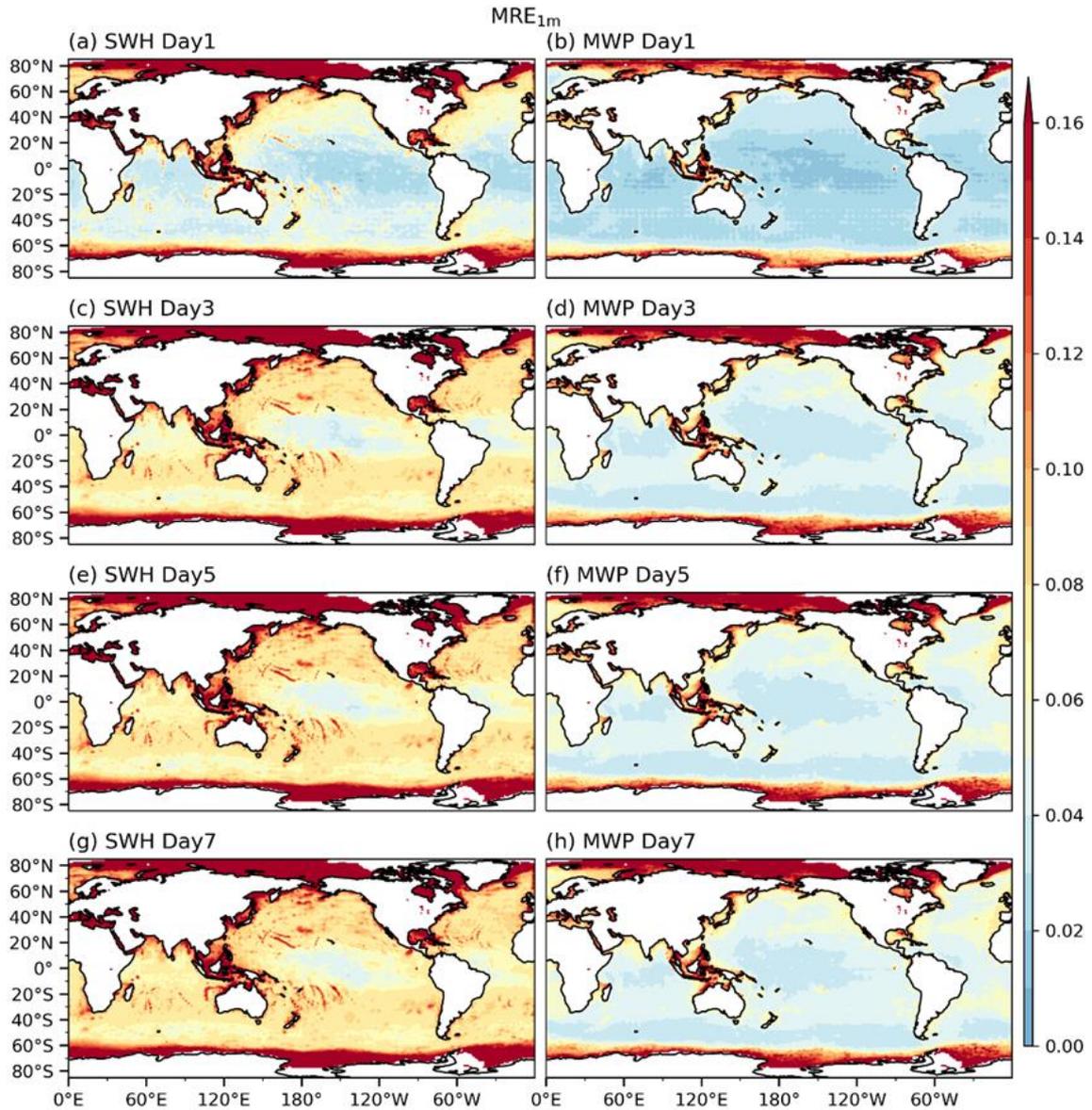

**Figure 4.** The $MRE_{1m}$ of (a, c, e, g) SWH, (b, d, f, h) MWP predicted by the Vision Transformer model at lead times of 1, 3, 5 and 7 days.

To further estimate the accuracy of the ViT model in predicting destructive waves, the global mean RMSE of the predicted waves at different heights of 1-8 meters are shown in Figure 5. For the predictions of the waves higher than 6 meters, the RMSE exceeds 1 meter. This indicates that the model cannot accurately predict the extreme ocean waves. The prediction error for the wave period is stable at around 1 second, and in the best case, the prediction error for 1-meter high waves at 1-day lead can be less than 0.8 seconds. Interestingly, the results suggest that it is relatively easier to accurately predict the direction of large ocean waves.

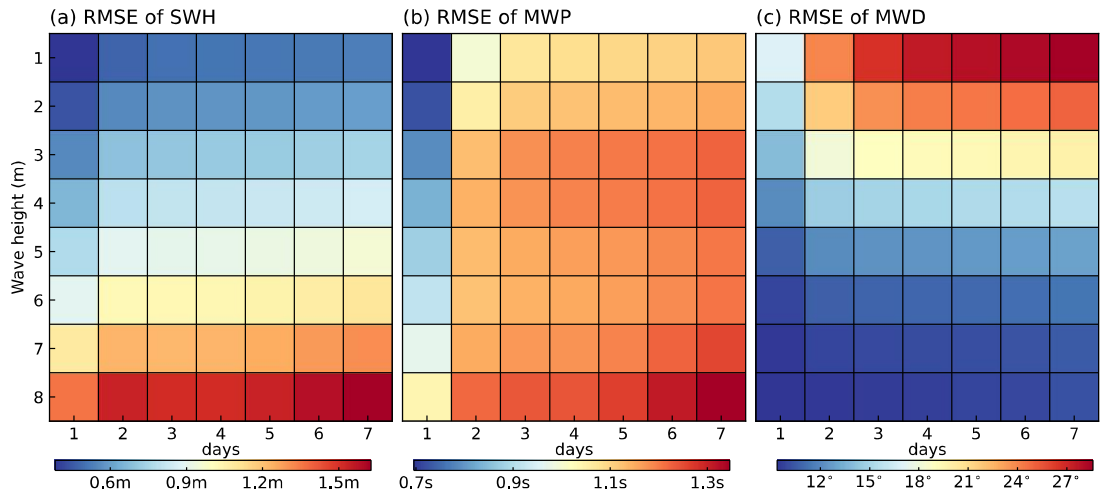

**Figure 5.** The global mean RMSE of (a) SWH, (b) MWP and (c) MWD for various wave heights (y-axis) predicted at lead times of 1-7 days(x-axis).

When a typhoon comes, it usually brings large waves. As a test case, Figure 6 shows the predicted and observed wave conditions in the northwestern Pacific caused by Typhoon "Muifa" in August 2011. The wave direction, wave period, and the center of maximum wave height are well predcited at 1-, 4- and 7-day lead, with an error of about 1 meter in predicting the maximum wave height. The result indicates that, giving an accurate wind field as input, the ViT model has a good predictive ability for strong waves caused by typhoons.

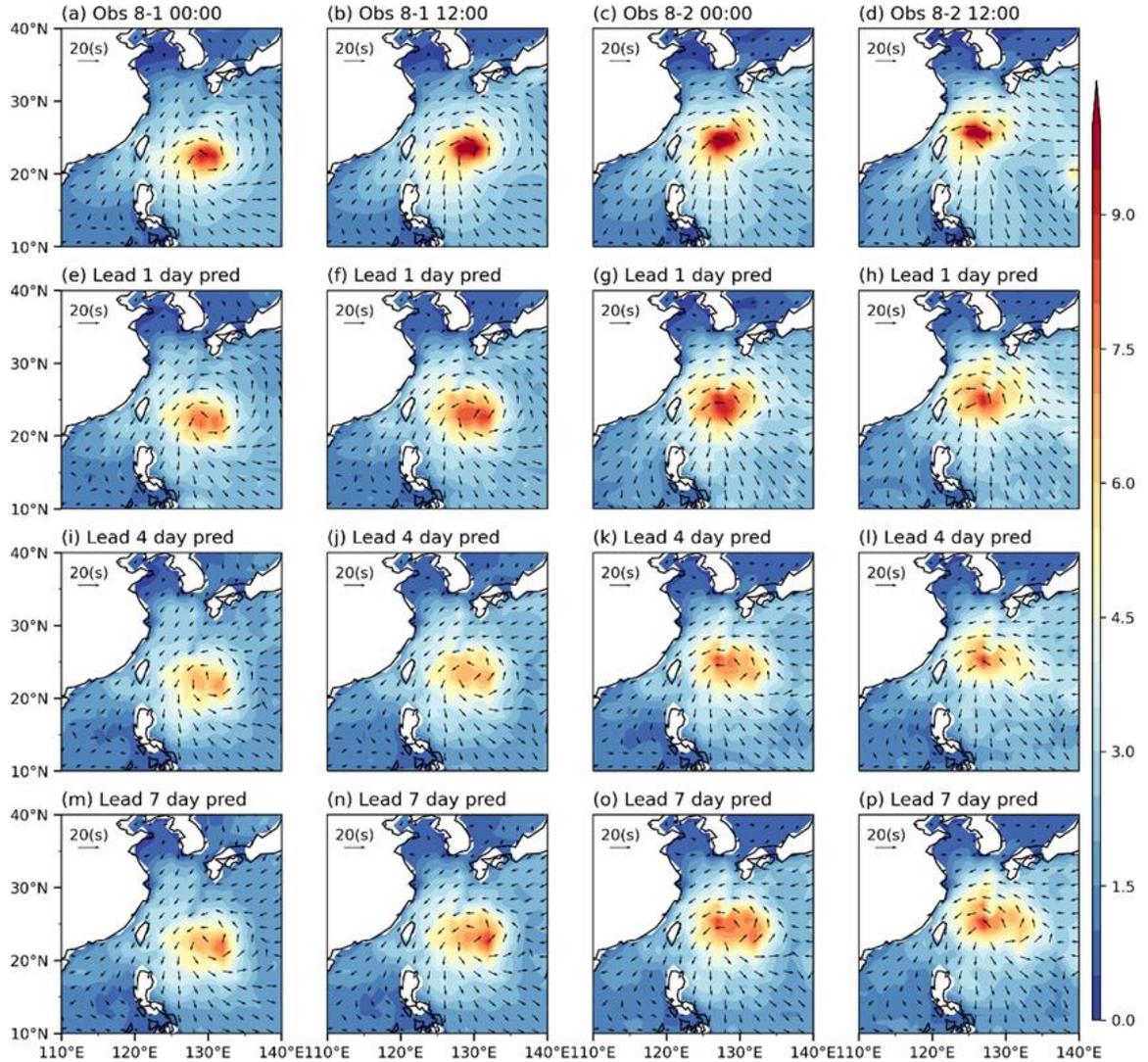

**Figure 6.** (a-d) Observed, (e-h)1- ,(i-l) 4- and (m-p) 7-day lead predicted waves associated with the approaching Typhoon "Muifa" in the northwestern Pacific (colored shading indicates SWH, arrows represents wave direction, and length represents wave period).

More evaluation of the ViT model performance was conducted (Figure 7). We first compare the prediction skills of the ViT and ConvLSTM models based on the test dataset. The result indicates that the transformer model not only can be easily trained but also produces better prediction accuracy compared to the traditional convolutional architecture. Note that for real-time ocean wave forecasts, the future winds are unavailable and have to be supplied with predictions. Therefore, we also compare the skills based on a real-time forecast experiment. Undoubtedly, with predicted winds by IFS or FengWu as the input, the accuracy of the ViT model's prediction will be lower than that with ERA5 winds as the input. However, there is no significant difference in the prediction accuracy during the first three days. This means that it is feasible to forecast ocean waves with the predicted winds at short-lead times. It is worth noting that, except the relatively larger errors in predicting high waves at 5-7 days, the prediction accuracy of the model with the wind input predicted by FengWu is comparable to that with the wind input predicted by IFS (Figure 7g-h).

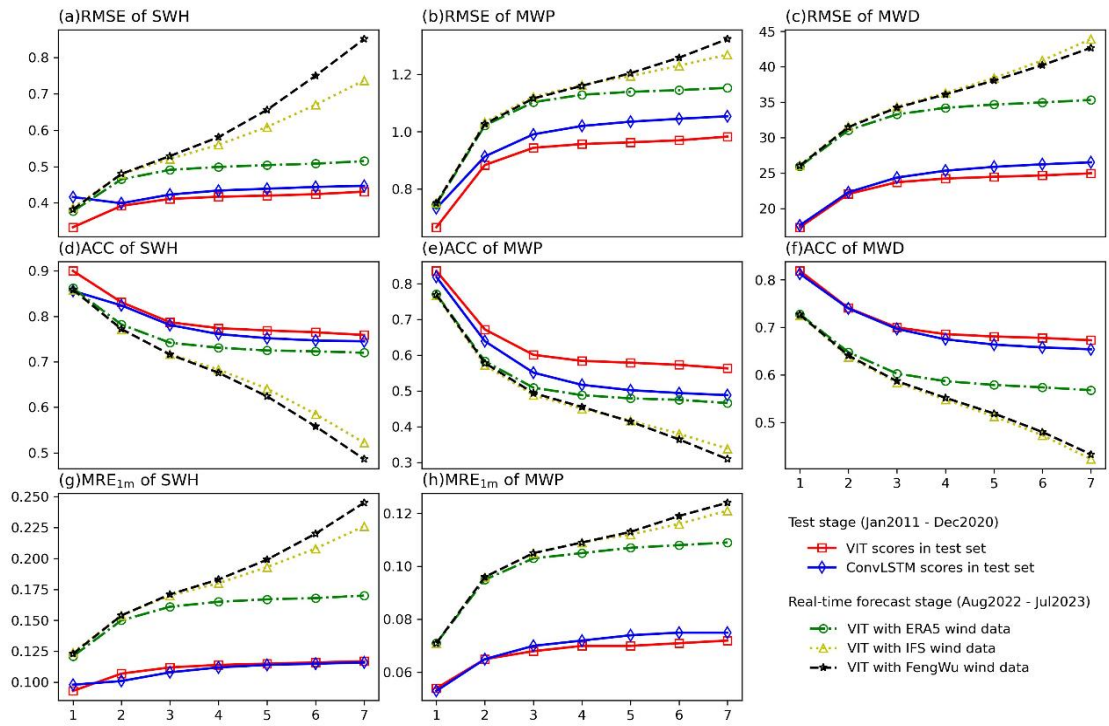

**Figure 7.** Skill comparison based on different models and wind inputs. The red (blue) curve represents the skills of the ViT (ConvLSTM) model in predicting global ocean waves at 1-7 days lead for the test period during Jan. 2011–Dec. 2020. For the real-time forecasts during Aug. 2022–Jul. 2023, the skill comparison is conducted based on the ViT model only, and the green (yellow/black) line represents the model's prediction skill with the ERA5 reanalysis (IFS/FengWu predicted) winds being adopted as the input.

There are many ways to build AI-based large models to generate and improve ocean wave forecasts. It is also possible to directly train wave forecast models based on the winds predicted by AI large models. Leveraging the ability of AI models to quickly and effectively perform predictions, wave forecasts based on AI large models have enormous potential and great application values to various sectors.

## 6. Summary

In this paper, we introduce the evolution of AI-based weather forecast models, highlighting the potential of AI models in developing comprehensive, data-driven weather forecast applications. Based on the common characteristics of influential large-parameter models in recent years, we conceptualize AI weather forecast models through the framework of the "Three Large Rules". We anticipate that, as AI models develop further, they will rapidly iterate and steer towards practical applications, and fundamentally revolutionizing our current weather forecast methods. This transformation promises enhanced accuracy, cost-effectiveness, and high computational efficiency. This shift brings forth numerous opportunities and challenges in the field of weather forecasting.

However, although LWM has the potential to revolutionize weather forecasting in terms of accuracy and efficiency, a balanced perspective is essential. The coexistence with traditional NWP models is critical, as they provide a nuanced understanding of the intricate physical processes underlying atmospheric phenomena. These dynamical models are invaluable, providing essential insights that complement the AI's data-centric approach. Moreover, our analysis of the current AI modeling solutions reveals a symbiotic relationship between AI and traditional NWP models. The latter continues to serve as the bedrock for the former, offering essential training data and serving as a benchmark for validation.

A pinnacle example of this harmonious integration is exemplified in the success of Neural GCM. This groundbreaking model illustrates that physics-based and AI models are not mutually exclusive entities but collaborative components. Their convergence enhances predictive capabilities and signifies a holistic approach for more accurately simulation and forecasts of the complex weather and climate. This integration, whether it is adding physical equations to AI models or leveraging AI methods in NWP, both promises to open a new avenue for improved understanding of atmospheric dynamics, leading to a more comprehensive and reliable framework for forecasting weather and climate changes (Figure 8).

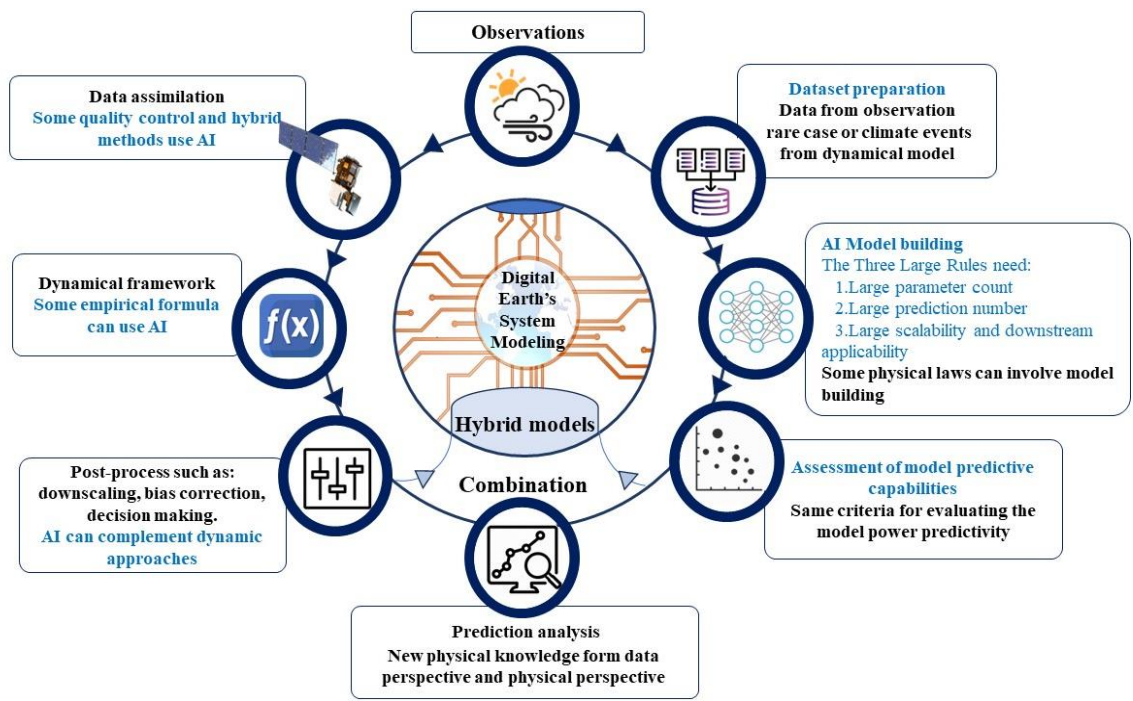

**Figure 8.** Schematic diagram of the possible integration of large AI models with dynamical models for improving ocean-atmosphere forecasts.

**Reference**

1. Abbe, C. 1901. The physical basis of long-range weather forecasts. Mon. Weather Rev., 29, 551–561.
2. Alley, R.B., Emanuel, K.A., Zhang, F. 2019. Advances in weather prediction. Science, 363, 342–344.
3. Andrychowicz, M., Espeholt, L., Li, D., Merchant, S., Merose, A., Zyda, F., Agrawal, S., Kalchbrenner, N., Deepmind, G., Research, G. 2023. Deep learning for day forecasts from sparse observations. arXiv preprint arXiv:2306.06079.
4. Azimi-Sadjadi, M.R., Zekavat, S.A. 2000. Cloud classification using support vector machines. In IGARSS 2000. IEEE 2000 International Geoscience and Remote Sensing Symposium. Taking the Pulse of the Planet: The Role of Remote Sensing in Managing the Environment. Proceedings (Cat. No.00CH37120), IEEE, pp. 669–671.
5. Bauer, P., Thorpe, A., Brunet, G. 2015. The quiet revolution of numerical weather prediction. Nature, 525(7567), 47–55.
6. Ben-Bouallegue, Z., Clare, M.C.A., Magnusson, L., Gascon, E., Maier-Gerber, M., Janousek, M., Rodwell, M., Pinault, F., Dramsch, J.S., Lang, S.T.K., Raoult, B., Rabier, F., Chevallier, M., Sandu, I., Dueben, P., Chantry, M., Pappenberger, F. 2023. The rise of data-driven weather forecasting. arXiv preprint arXiv:2307.10128.
7. Benjamin, S.G., Brown, J.M., Brunet G., Lynch, P., Saito, K., Schlatter, T.W. 2018. 100 years of progress in forecasting and NWP applications. Meteor. Monogr., 59, 13.1-13.67.
8. Bi, K., Xie, L., Zhang, H., Chen, X., Gu, X., Tian, Q. 2023. Accurate medium-range global weather forecasting with 3D neural networks. Nature, 619, 533–538.
9. Bjerknes, V. 1904. Das problem der wettervorhersage, betrachtet vom Standpunkte der Mechanik und der Physik. Meteor. Z., 1–7.
10. Charney, J.G., Fjörtoft, R., Neumann, & J.V. Von. 1950a. Numerical integration of the

barotropic vorticity equation. Tellus., 2, 237–254.
11. Chen, Kang, Han, T., Gong, J., Bai, L., Ling, F., Luo, J.-J., Chen, X., Ma, L., Zhang, T., Su, R., Ci, Y., Li, B., Yang, X., Ouyang, W., 2023. FengWu: Pushing the skillful global medium-range weather forecast beyond 10 days lead. arXiv preprint arXiv:2304.02948.
12. Chen, Kun, Bai, L., Ling, F., Ye, P., Chen, T., Chen, K., Han, T., Ouyang, W. 2023. Towards an end-to-end artificial intelligence driven global weather forecasting system. arXiv preprint arXiv:2312.12462.
13. Chen, L., Zhong, X., Wu, J., Chen, D., Xie, S., Chao, Q., Lin, C., Hu, Z., Lu, B., Li, H., Qi, Y. 2023b. FuXi-S2S: an accurate machine learning model for global subseasonal forecasts. arXiv preprint arXiv:2312.09926.
14. Chen, L., Zhong, X., Zhang, F., Cheng, Y., Xu, Y., Qi, Y., Li, H. 2023a. FuXi: a cascade machine learning forecasting system for 15-day global weather forecast. npj Clim. Atmos. Sci., 6, 190.
15. Courtier, P., Thépaut, J. -N, Hollingsworth, A. 1994. A strategy for operational implementation of 4D-Var, using an incremental approach. Q. J. ROY. METEOR. SOC., 120, 1367–1387.
16. Dosovitskiy, A., Beyer, L., Kolesnikov, A., Weissenborn, D., Zhai, X., Unterthiner, T., Dehghani, M., Minderer, M., Heigold, G., Gelly, S., Uszkoreit, J., Houlsby, N. 2020. An Image is Worth 16x16 Words: Transformers for Image Recognition at Scale. International Conference on Learning Representations.
17. Dueben P D, Bauer P. Challenges and design choices for global weather and climate models based on machine learning. Geo. Model Devel., 11(10): 3999-4009.
18. Dueben, P.D., Bauer, P. 2018. Challenges and design choices for global weather and climate models based on machine learning. GEOSCI. MODEL DEV., 11, 3999–4009.
19. Guibas, J., Mardani, M., Li, Z., Tao, A., Aanandkumar, A., Catanzaro, B. 2021. Adaptive fourier neural operators: efficient token mixers for transformers. International Conference on Learning Representations.
20. Hakim, G.J., Masanam, S. 2023. Dynamical Tests of a Deep-Learning Weather Prediction Model. arXiv preprint arXiv:2309.10867.
21. Han T, Guo S, Ling F, Chen K., Gong J., Luo, J.-J., Gu J., Dai K., Ouyang, W., Bai, L. 2024. FengWu-GHR: learning the kilometer-scale mmedium-range global weather forecasting. arXiv preprint arXiv:2402.00059.
22. Hess, P., Drüke, M., Petri, S., Strnad, F.M., Boers, N. 2022. Physically constrained generative adversarial networks for improving precipitation fields from Earth system models. NAT. MACH. INTELL., 4, 828–839.
23. Hsieh, W., Tang B., 1998. Applying neural network models to prediction and data analysis in meteorology and oceanography. B. AM. METEOROL. SOC., 1855–1870.
24. Hu, Y., Chen, L., Wang, Z., Li, H. 2023. SwinVRNN: a data-driven ensemble forecasting model via learned distribution perturbation. J. ADV. MODEL EARTH SY., 15, e2022MS003211.
25. Jumper, J., Evans, R., Pritzel, A., Green, T., Figurnov, M., Ronneberger, O., Tunyasuvunakool, K., Bates, R., Žídek, A., Potapenko, A., Bridgland, A., Meyer, C., Kohl, S.A.A., Ballard, A.J., Cowie, A., Romera-Paredes, B., Nikolov, S., Jain, R., Adler, J., Back, T., Petersen, S., Reiman, D., Clancy, E., Zielinski, M., Steinegger, M., Pacholska, M.,

Berghammer, T., Bodenstein, S., Silver, D., Vinyals, O., Senior, A.W., Kavukcuoglu, K., Kohli, P., Hassabis, D. 2021. Highly accurate protein structure prediction with AlphaFold. Nature, 596, 583–589.

26. Keisler, R. 2022. Forecasting global weather with graph neural networks. arXiv preprint arXiv:2202.07575.

27. Kochkov, D., Yuval, J., Langmore, I., Norgaard, P., Smith, J., Mooers, G., Lottes, J., Rasp, S., Düben, P., Klöwer, M., Hatfield, S., Battaglia, P., Sanchez-Gonzalez, A., Willson, M., Brenner, M.P., Hoyer, S. 2023. Neural General Circulation Models. arXiv preprint arXiv:2311.07222.

28. Lam, R., Sanchez-Gonzalez, A., Willson, M., Wirnsberger, P., Fortunato, M., Alet, F., Ravuri, S., Ewalds, T., Eaton-Rosen, Z., Hu, W., Merose, A., Hoyer, S., Holland, G., Vinyals, O., Stott, J., Pritzel, A., Mohamed, S., Battaglia, P. 2023. Learning skillful medium-range global weather forecasting. Science, eadi2336.

29. Lazo, J.K., Morss, R.E., Demuth, J.L. 2009. 300 billion served: sources, perceptions, uses, and values of weather forecasts. B. AM. METEOROL. SOC., 90, 785–798.

30. Li, W., Liu, Z., Chen, K., Chen, H., Liang, S., Zou, Z., Shi, Z. 2024. DeepPhysiNet: Bridging Deep Learning and Atmospheric Physics for Accurate and Continuous Weather Modeling. arXiv preprint arXiv:2401.04125.

31. Ling, F., Li, Y., Luo, J.J., Zhong, X., Wang, Z. 2022. Two deep learning-based bias-correction pathways improve summer precipitation prediction over China. ENVIRON. RES. LETT., 17, 124025.

32. Ling, F., Lu, Z., Luo, J.-J., Bai, L., Behera, S. K., Jin, D., Pan, B., Jiang, H., Yamagata, T. 2024. Diffusion Model-based Probabilistic Downscaling for 180-year East Asian Climate Reconstruction. arXiv preprint arXiv:2402.06646.

33. Liu, Z., Chen, H., Bai, L., Li, W., Chen, K., Wang, Z., Ouyang, W., Zou, Z., Shi, Z. (2024). Observation-Guided Meteorological Field Downscaling at Station Scale: A Benchmark and a New Method. arXiv preprint arXiv:2401.11960.

34. Liu, Z., Lin, Y., Cao, Y., Hu, H., Wei, Y., Zhang, Z., Lin, S., Guo, B. 2021. Swin transformer: Hierarchical vision transformer using shifted windows. Proceedings of the IEEE/CVF international conference on computer vision, 10012-10022.

35. Lynch, P. 2008. The origins of computer weather prediction and climate modeling. J. COMPUT. PHYS., 227, 3431–3444.

36. Melinc, B., Zaplotnik, Ž. 2023. Neural-network data assimilation using variational autoencoder. arXiv preprint arXiv:2308.16073.

37. Neukom, R., Barboza, L.A., Erb, M.P., Shi, F., Emile-Geay, J., Evans, M.N., Franke, J., Kaufman, D.S., Lücke, L., Rehfeld, K., Schurer, A., Zhu, F., Brönnimann, S., Hakim, G.J., Henley, B.J., Ljungqvist, F.C., McKay, N., Valler, V., von Gunten, L. 2019. Consistent multidecadal variability in global temperature reconstructions and simulations over the Common Era. Nat. Geosci., 12, 643–649.

38. Nguyen, T., Brandstetter, J., Kapoor, A., Gupta, J.K., Grover, A. 2023. ClimaX: a foundation model for weather and climate. arXiv preprint arXiv:2301.10343.

39. P. Bauer, et al., 2020, The ECMWF scalability programme: Progress and plans European Centre for Medium Range Weather Forecasts.

40. Pan, B., Wang, L. Y., Zhang, F., Duan, Q., Li, X., Pan, X., Chen, X., Ling F., Wang S., Pan,

M., Xiao, Z. 2023 Probabilistic diffusion model for stochastic parameterization--a case example of numerical precipitation estimation. Authorea Preprints.

41. Pathak, J., Subramanian, S., Harrington, P., Raja, S., Chattopadhyay, A., Mardani, M., Kurth, T., Hall, D., Li, Z., Azizzadenesheli, K., Hassanzadeh, P., Kashinath, K., Anandkumar, A. 2022. FourCastNet: a global data-driven high-resolution weather model using adaptive fourier neural operators. arXiv preprint arXiv:2202.11214.

42. Pfaff, T., Fortunato, M., Sanchez-Gonzalez, A., Battaglia, P.W. 2020. Learning mesh-based simulation with graph networks. International Conference on Learning Representations.

43. Price, I., Sanchez-Gonzalez, A., Alet, F., Ewalds, T., El-Kadi, A., Stott, J., Mohamed, S., Battaglia, P., Lam, R., Willson, M., Deepmind, G. 2023. GenCast: Diffusion-based ensemble forecasting for medium-range weather. arXiv preprint arXiv:2312.15796.

44. Rahmstorf, S., Coumou, D. 2011. Increase of extreme events in a warming world. Proc. Natl. Acad. Sci. USA, 108, 17905–17909.

45. Rasp S, Thuerey N. Data-driven medium-range weather prediction with a resnet pretrained on climate simulations: A new model for weatherbench. J. ADV. MODEL EARTH SY., 13(2): e2020MS002405.

46. Rasp, S., Dueben, P.D., Scher, S., Weyn, J.A., Mouatadid, S., Thuerey, N. 2020. WeatherBench: a benchmark data set for data-driven weather forecasting. J. ADV. MODEL EARTH SY., 12, e2020MS002203.

47. Rasp, S., Hoyer, S., Merose, A., Langmore, I., Battaglia, P., Russell, T., Sanchez-Gonzalez, A., Yang, V., Carver, R., Agrawal, S., Chantry, M., Bouallegue, Z. Ben, Dueben, P., Bromberg, C., Sisk, J., Barrington, L., Bell, A., Sha, F. 2023. WeatherBench 2: a benchmark for the next generation of data-driven global weather models. arXiv preprint arXiv:2308.15560.

48. Schaul, T., Quan, J., Antonoglou, I., Silver, D. 2015. Prioritized experience replay. International Conference on Learning Representations.

49. Scher, S. 2018. Toward data-driven weather and climate forecasting: approximating a simple general circulation model with deep learning. Geophys. Res. Lett., 45, 12,616-12,622.

50. Schultz, M.G., Betancourt, C., Gong, B., Kleinert, F., Langguth, M., Leufen, L.H., Mozaffari, A., Stadtler, S. 2021. Can deep learning beat numerical weather prediction? Phil. Trans. R. Soc. A., 379, 20200097.

51. Selz, T., Craig, G.C. 2023. Can artificial intelligence-based weather prediction models simulate the butterfly effect? Geophys. Res. Lett., 50, e2023GL105747.

52. Silver, D., Schrittwieser, J., Simonyan, K., Antonoglou, I., Huang, A., Guez, A., Hubert, T., Baker, L., Lai, M., Bolton, A., Chen, Y., Lillicrap, T., Hui, F., Sifre, L., Van Den Driessche, G., Graepel, T., Hassabis, D. 2017. Mastering the game of Go without human knowledge. Nature, 550, 354–359.

53. Stensrud, D. 2009. Parameterization schemes: keys to understanding numerical weather prediction models. Cambridge University Press.

54. Tsagkatakis, G., Aidini, A., Fotiadou, K., Giannopoulos, M., Pentari, A., Tsakalides, P. 2019. Survey of deep-learning approaches for remote sensing observation enhancement Sensors, 19, 3929.

55. Vaswani, A., Brain, G., Shazeer, N., Parmar, N., Uszkoreit, J., Jones, L., Gomez, A.N.,

Kaiser, Ł., Polosukhin, I. 2017. Attention is all you need. Advances in neural information processing systems, 30.
56. Wang, Y., Shi, X., Lei, L., Fung, J.C.H. 2022. Deep learning augmented data assimilation: reconstructing missing information with convolutional autoencoders. Mon. Weather Rev., 150, 1977–1991.
57. Watt-Meyer, O., Dresdner, G., McGibbon, J., Clark, S.K., Henn, B., Duncan, J., Brenowitz, N.D., Kashinath, K., Pritchard, M.S., Bonev, B., Peters, M.E., Bretherton, C.S. 2023. ACE: A fast, skillful learned global atmospheric model for climate prediction. arXiv preprint arXiv:2310.02074.
58. Weyn, J.A., Durran, D.R., Caruana, R. 2019. Can machines learn to predict weather? using deep learning to predict gridded 500-hPa geopotential height from historical weather data. J. ADV. MODEL EARTH SY., 11, 2680–2693.
59. Williams, P.D. 2005. Modelling climate change: the role of unresolved processes. Phil. Trans. R. Soc. *A*. 363, 2931–2946.
60. Wu, H., Zhou, H., Long, M., Wang, J. 2023. Interpretable weather forecasting for worldwide stations with a unified deep model. NAT. MACH. INTELL., 5, 602–611.
61. Xiao, Y., Bai, L., Xue, W., Chen, K., Han, T., Ouyang, W. 2023. FengWu-4DVar: Coupling the Data-driven Weather Forecasting Model with 4D Variational Assimilation. arXiv preprint arXiv:2312.12455.
62. Zhang, Y., Long, M., Chen, K., Xing, L., Jin, R., Jordan, M.I., Wang, J. 2023. Skilful nowcasting of extreme precipitation with NowcastNet. Nature, 619, 526–532.
63. Zhong, X., Chen, L., Liu, J., Lin, C., Qi, Y., Li, H. 2023. FuXi-Extreme: Improving extreme rainfall and wind forecasts with diffusion model. arXiv preprint arXiv:2310.19822

**Acknowledgments**

This work is supported by the National Key Research and Development Program of China (No. 2020YFA0608000) and National Natural Science Foundation of China (Grant 42030605).

**Author contributions**

F.L., L.O. and B.R.L. are co-first authors. F.L., L.O. and B.R.L. wrote the initial manuscript. L.O. designed the AI models and performed the main experiments under supervision of F.H.L. and L.B. J.-J.L. provided the idea, revised the manuscript and supervised the entire work. All authors contributed to writing, interpreting results and discussions.

**COMPETING INTERESTS**

All authors declare no competing interests.